\pdfoutput=1

\documentclass[11pt]{article}

\usepackage{EMNLP2022}

\usepackage{times}
\usepackage{latexsym}

\usepackage[T1]{fontenc}

\usepackage[utf8]{inputenc}

\usepackage{microtype}

\usepackage{subfiles}

\usepackage{booktabs}
\usepackage{multirow}
\usepackage{stfloats}
\usepackage{lipsum}
\usepackage{arydshln}

\usepackage{graphicx}

\usepackage[nointegrals]{wasysym}
\usepackage{amsthm}
\usepackage{amssymb}
\usepackage{amsmath}
\usepackage[normalem]{ulem}
\usepackage{color} 
\usepackage{enumitem}
\usepackage{mathtools}

\newcommand{\vic}[1]{\textcolor{black}{#1}}
\newcommand{\vicf}[1]{\textcolor{black}{#1}}

\title{Transformer-based Entity Typing in Knowledge Graphs}
\author{Zhiwei Hu$^\clubsuit$ \quad V\'ictor Guti\'errez-Basulto$^\diamondsuit$ \quad  Zhiliang Xiang$^\diamondsuit$  \\ {\bf Ru Li}$^{\clubsuit *}$ \qquad {\bf Jeff Z. Pan}$^{\spadesuit}$\thanks{ \, Contact Authors} \\
$\clubsuit$ School of Computer and Information Technology, Shanxi University, China\\
$\diamondsuit$ School of Computer Science and Informatics, Cardiff University, UK\\
$\spadesuit$ ILCC, School of Informatics, University of Edinburgh, UK \\
$\clubsuit$ \texttt{zhiweihu@whu.edu.cn,liru@sxu.edu.cn}\\
$\diamondsuit$\texttt{\{gutierrezbasultov,xiangz6\}@cardiff.ac.uk}\\
$\spadesuit$\texttt{j.z.pan@ed.ac.uk}\\} 

\begin{document}
\maketitle

\begin{abstract}
\vic{We investigate the knowledge graph entity typing task  which  aims at inferring plausible entity types.} 
In this paper, we propose  a novel \textbf{T}ransformer-based \textbf{E}ntity \textbf{T}yping  (TET) approach, \vicf{effectively encoding the content of  neighbors of an entity}. More precisely, TET is composed of three different mechanisms:  a \textbf{\textit{local transformer}} allowing to infer  missing  types \vicf{of an enity} by independently encoding the information provided by each of its neighbors; a \textbf{\textit{global transformer}} aggregating the information of all neighbors of an entity into a single long sequence  to
reason about more complex entity types; and a \textbf{\textit{context transformer}} integrating neighbors content \vicf{based on their contribution to the type inference} through information exchange between neighbor pairs. Furthermore, TET uses  information about class membership of types  to semantically strengthen the representation of an entity.   Experiments on two real-world datasets demonstrate the superior performance of TET compared to the state-of-the-art.   

\end{abstract}

\section{Introduction}
\vic{A knowledge graph (KG)~\cite{Pan2016} is  a multi-relational graph encoding factual knowledge, 
with the form $(h, r, t)$ where $h$, $t$ are the head and tail entities connected via  the relation   $r$. In this paper, we consider KGs with \emph{minimal schema information}, i.e.,  those    containing  entity type assertions, as the only schema information,  of the form $(e,\textit{has\_type},c)$ stating that the entity $e$ has type $c$; \vicf{e.g., to capture that Barack Obama has type President}. Entity type knowledge is widely used in  NLP  tasks, e.g., in  relation extraction \citep{Liu_2014}, entity  and    relation    linking \citep{Gupta_2017,PZSv+2019}, question answering \citep{ElSahar_2018, Hu_2022}, and fine-grained entity typing on text \citep{Onoe_2021, Qian_2021, Liu_2021}. However, entity types are far from complete, since in real-world applications they are continuously emerging.  For example, about 10\% of entities in FB15k \citep{Bordes_2013} have the type \textit{/music/artist}, but do not have \textit{/people/person}~\citep{Moon_2017}. }

\begin{figure}[t!]
    \centering
    \includegraphics[width=0.43\textwidth]{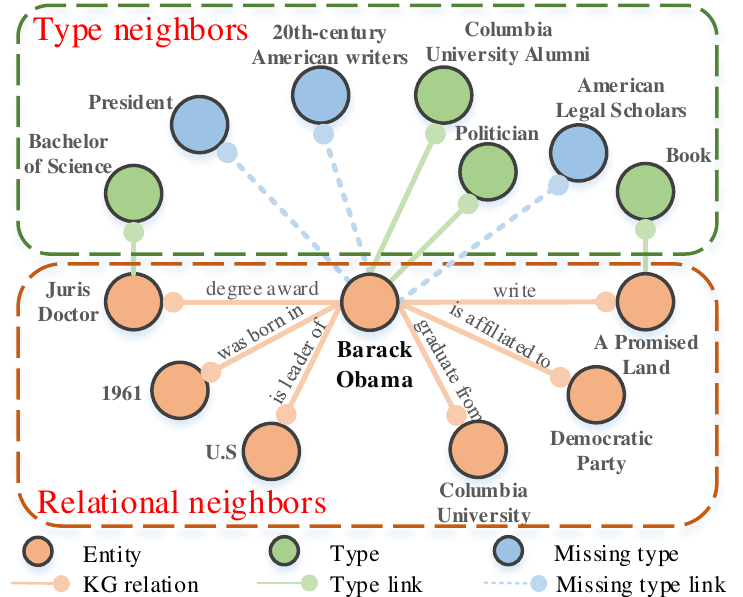}
    \caption{A KG with its entity type information.}
    \label{figure_instance}
\end{figure}

\vic{In light of this, it has been recently investigated the \emph{Knowledge Graph Entity Typing (KGET)} task, aiming at inferring missing entity types in a KG.} 
Most existing approaches to KGET use methods based on either embeddings or graph convolutional networks (GCN). Despite the huge progress these methods have made, there are still some  important challenges to be solved. \vicf{On the one hand,} most embedding-based  models~\cite{Moon_2017,Zhao_2020,Ge_2021,Zhuo_2022} encode all neighbors of a target entity into a single vector, but in many cases only some neighbors are necessary to infer the correct types. For example, as shown in Figure~\ref{figure_instance}, to predict that the entity \textit{Barack Obama} has type \textit{President}, only the neighbor $\xrightarrow{\textit{is\_leader\_of}}$ \textit{U.S} 
%
is needed. Indeed, using too many neighbors, such as $\xrightarrow{\textit{graduate\_from}}$ \textit{Columbia University},
%
will 
introduce noise. The CET model~ \citep{Pan_2021} overcomes this problem  by encoding each neighbor independently. However, since entities and relations are represented by TransE \citep{Bordes_2013}, \vicf{there is a restriction on the direction of the representation of entities and relations direction, fixing it from entity to relation or vice versa. As a consequence, \textbf{\textit{certain interactions between neighbor entities and relations are ignored}}}. Also, to  predict more complex types, CET directly adds and averages the neighbor representations, \textbf{\textit{ weakening the contribution of different neighbors}}, \vicf{since it ignores that the contribution of different neighbors to different types might not be the same}. For example, as shown in Figure~\ref{figure_instance}, the inference of the type  \textit{20th-century American writer} involves multiple semantic aspects of  \textit{Barack Obama}, it requires to jointly consider the neighbors
$\xrightarrow{\textit{write}}$ \textit{A Promised Land}, $\xrightarrow{\textit{was\_born\_in}}$ \textit{1961}, and $\xrightarrow{\textit{is\_leader\_of}}$  \textit{U.S}, but the neighbor
$\xrightarrow{\textit{degree\_award}}$ \textit{Juris Doctor} should get less attention.
\vicf{On the other hand,} \vic{GCN   frameworks for KGET use expressive representations for entities and relations based on their neighbor entities and relations~\citep{Jin_2019,Zhao_2022,Zou_2022,Vashishth_2020,Pan_2021}. However, a common  problem of GCN-based models is that they aggregate information only along the paths starting from neighbors of the target entity, \textbf{\textit{limiting the representation of   interdependence between  neighbors that are not directly connected.}} For example, in Figure~\ref{figure_instance} the  entities \textit{Juris Doctor} and \textit{U.S} are not connected, but combining their information could help to infer that  \textit{American Legal Scholars} is a type  of  \textit{Barack Obama}. 
This could   be fixed by increasing the number of  layers, but with an additional computational cost.}

\vic{The main objective of this paper is to introduce a transformer-based approach to KGET that addresses  the highlighted challenges. The transformer architecture~\citep{Vaswani_2017} has been essential for NLP, e.g., in  pre-trained language models  \citep{Devlin_2019, Reimers_2019, Lan_2020, Wu_2021b}, document modeling \citep{Wu_2021a}, and link prediction \citep{Wang_2019, Chen_2021}. Transformers are well-suited for KGET as entities and relations in a KG can be regarded as tokens, and using the transformer as encoder, one can thus achieve bidirectional deep interaction between entities and relations. Specifically, we propose \textbf{TET}, a \textbf{T}ransformer-based \textbf{E}ntity \textbf{T}yping  model for KGET, composed of the following three  inference modules. A \textbf{\textit{local transformer}} that independently  encodes the \vicf{relational and type}  neighbors of an entity into a sequence, facilitating bidirectional interaction between elements within the sequence, addressing the first problem. A \textbf{\textit{global transformer}} that aggregates all neighbors of an  entity into a single long sequence to simultaneously consider multiple attributes of an entity, allowing to infer more `complex' types,  thus addressing  the third problem. A \textbf{\textit{context transformer}} that   aggregates  neighbors of an entity  in a differentiated manner \vicf{according to their contribution} while preserving the graph structure,   thus addressing  the second problem. Furthermore, we use semantic knowledge about the known types in a KG. In particular, we find out that types are normally clustered in classes. For example, the  types \textit{medicine/disease}, \textit{medicine/symptom}, and \textit{medicine/drug}  belong to the class  \textit{medicine}. We use this class membership information for replacing the \vicf{`generic'} relation \textit{has\_type}  \vicf{with a more fine-grained relation that captures to which class a type belongs to},  enriching  the semantic content of connections between entities and types.} \vic{ To sum up, our contributions are:}

\begin{itemize}[itemsep=1ex, leftmargin=5mm]
  \item \vic{We propose a novel transformer-based framework for inferring missing entity types  in KGs, encoding knowledge about entity neighbors from three different perspectives.} 
  \item \vic{We use class membership of types to replace the single \textit{has\_type} relation \vicf{with  class-membership relations providing fine-grained semantic information}.}
\item\vic{ We conduct empirical and ablation experiments on two real-world datasets,  demonstrating the superiority of TET over existing SoTA models.}
\end{itemize}

Data, code, and an extended version with appendix are available at \url{https://github.com/zhiweihu1103/ET-TET}.


\section{Related Work}
\vic{The  knowledge graph completion (KGC)  task is usually concerned with predicting   the missing head or tail entities of  a triple. 
KGET can thus be seen as a specialization of   KGC. Existing KGET methods can be \vicf{classified} in embedding- and GNC-based.} 
\paragraph{Embedding-based Methods.} 
\vic{ETE \citep{Moon_2017} learns  entity embeddings for KGs by a standard representation learning method \citep{Bordes_2013}, and further builds a mechanism for information exchange between entities and their types. ConnectE \citep{Zhao_2020} jointly embeds entities and types into two different spaces and learns a mapping from the entity space to the type space. CORE~\citep{Ge_2021} utilizes the  models RotatE~\citep{Sun_2019} and ComplEx~\citep{Trouillon_2016} to embed entities and types into two different complex spaces, and develops a  regression model to link them. However,  the above methods do not fully consider the known types of entities while training the entity embedding representation, which seriously affects the prediction performance of missing types. Also, the representation of types in these methods is such that they cannot be semantically differentiated. CET~\citep{Pan_2021} jointly utilizes information about existing \vicf{type assertions} in a KG and about the neighborhood  of entities by respectively employing an independent-based mechanism and an aggregated-based one. It also utilizes a  pooling method to aggregate \vicf{their inference results}. AttEt \citep{Zhuo_2022} designs an attention mechanism to aggregate the neighborhood knowledge of an entity using type-specific weights, which are beneficial to capture specific characteristics of different types. A shortcoming of these two methods is that, unlike our TET model, they are not able to cluster types in classes, and are thus not able to semantically differentiate them in a  fine-grained way.} 
\paragraph{GCN-based Methods.} \vic{Graph Convolutional Networks (GCNs) have proven effective on modeling graph structures \citep{Kipf_2017, Hamilton_2017, Dettmers_2018}. However, directly using  GCNs on KGs usually leads to poor performance since KGs have different kinds of entities and relations. To address this problem, RGCN \citep{Schlichtkrull_2018} proposes to apply relation-specific transformations in GCN's aggregation. HMGCN \citep{Jin_2019} proposes a hierarchical multi-graph convolutional network to embed multiple kinds of semantic correlations between entities. CompGCN \citep{Vashishth_2020} uses composition operators from KG-embedding methods by jointly embedding both entities and relations in a relational graph. ConnectE-MRGAT \cite{Zhao_2022} proposes a multiplex relational graph attention network to learn on heterogeneous relational graphs, and then utilizes the ConnectE method for infering entity types. RACE2T \citep{Zou_2022} introduces a relational  graph attention network  method, utilizing the neighborhood  and relation information of an entity for type inference. A common problem with  these methods is that they follow a simple single-layer attention formulation, restricting the information transfer between unconnected neighbors of an entity. 
}

\paragraph{Transformer-based Methods.} To the best of our knowledge, there are no transformer-based approaches to KGET. However, two  transformer-based frameworks for the KGC task have been already proposed: CoKE~\cite{Wang_2019} and HittER~\cite{Chen_2021}. 
Our experiments show that they  are not suitable for  KGET. 

\section{Method}
\begin{figure*}[!htp]
    \centering
    \includegraphics[width=0.95\textwidth]{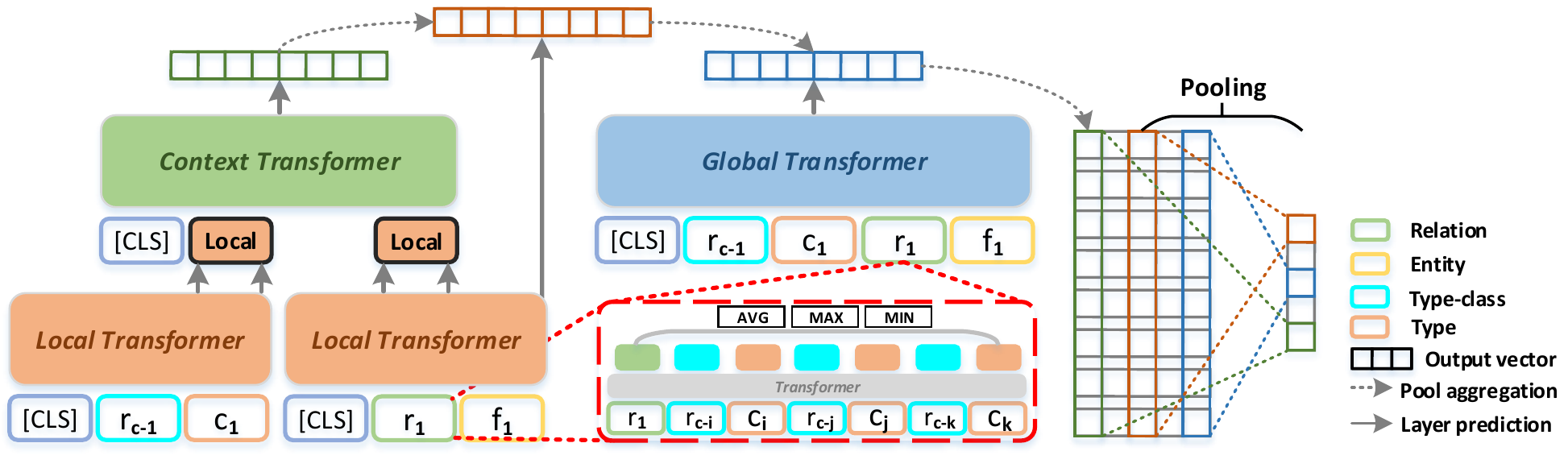}
    \caption{An overview of the TET model. The red dotted box part is only performed on the YAGO43kET dataset.  Note that $r_{c-i}$ is an abbreviation of $r_{class_{i}}$. Box with Local text indicates the output of the local transformer module.}
    \label{figure_model_structure}
\end{figure*}

\vic{In this section, we describe the architecture of our TET model (cf.\ Figure~\ref{figure_model_structure}). We start by introducing necessary background (Sec.~\ref{section_3_1}),   then present in detail the architecture of TET (Sec.~\ref{section_3_2}). Finally, we describe  pooling and optimization strategies (Sec.~\ref{section_3_3} and~\ref{section_3_4}).} 


\subsection{Background}
\label{section_3_1}
In this paper, a knowledge graph~\cite{Pan2016} is represented in a standard format for graph-structured data such as RDF~\cite{Pan09}.
A  \emph{knowledge graph (KG)} $\mathcal{G}$  is a tuple $(\mathcal{E},\mathcal{R},\mathcal C, \mathcal{T} )$, where  $\mathcal{E}$ is a set of entities, $\mathcal C$ is a set of entity types, $\mathcal{R}$ is a set of relation types,  and  $\mathcal{T}$  is a set of triples.
Triples in $\mathcal T$ are either
%
\emph{relation assertions}   ($h,r,t$)
where $h,t \in \mathcal{E}$  are respectively the \emph{head} and \emph{tail} entities of the triple, and $r \in \mathcal{R}$ is the \emph{edge}  of the triple connecting  head and tail; or  
%
\emph{entity type assertions} $(e,\textit{has\_type},c)$, where $e \in \mathcal E$, $c \in \mathcal C$, and \textit{has\_type} is the instance-of relation. 
For  $e \in \mathcal E$,   the \emph{relational neighbors of $e$} is the set  $\{(r,f) \mid (e,r,f) \in \mathcal T\}$. The \emph{type neighbors of $e$} are defined as $\{(\textit{has\_type},c) \mid (e,\textit{has\_type},c) \in \mathcal T\}$. We will simply say  \emph{neighbors of $ e$} when we refer to the  relational and type neighbors of $e$. \vic{ The goal of this paper is to address   KGET task which aims at inferring missing types from  $\mathcal C$ in entity type assertions.}

\subsection{Model Architecture}
\label{section_3_2}
\vic{In this section, we introduce the local, global and context  transformer-based modeling components of our TET model.} \vicf{Before defining these  components, we start by discussing an important observation.}

\subsubsection{Class Membership}
\vicf{
 A key observation is that in a KG \emph{all} type assertions are  uniformly defined using  the relation \textit{has\_type}. As a consequence, we do not have a way to fully differentiate the contribution of different types of an entity during inference, as we cannot capture the relationship between them and their relevance, weakening thus the contribution of type supervision on entities.
%
%
However, in practice types are clustered together in classes (i.e., root types in a domain); e.g., the types \textit{medicine/disease}, \textit{medicine/symptom}, and \textit{medicine/drug}  belong to the class \textit{medicine}. This  allows us to identify that  these types are related as all of them talk about something related to medicine, providing  us therefore with   fine-grained semantic information.  With this insight in mind, for each class, we create a relation  that will be used to model that a type is an element of that class. For instance, for the class \textit{medicine}, we introduce the relation \textit{belongs\_class\_medicine} . We will then replace a type neighbor $(\textit{has\_type},c)$ of an entity $e$ with $(r_{\textit{class}}, c)$, where $r_{\textit{class}}$ is the relation modeling  class membership, i.e. belonging to the class medicine. 
We define the \emph{type-class neighbors of an entity} as expected. We will use below this semantically-enriched representation in our local and global transformers.}
\subsubsection{Local Transformer}\label{sec:local} \vic{The main intuition behind the local component is that the neighbors of an entity might help to determine its types, and that the contribution of each neighbor  is different.
%
For instance, if  the entity \textit{Liverpool} has the relational neighbor (\textit{places\_lived}, \textit{Daniel Craig}), it is plausible to infer \textit{Liverpool} has type \textit{/location/citytown}. On the other hand, the neighbor (\textit{sports\_team}, \textit{Liverpool F.C.}) may help to infer that it has type \textit{/sports/sports\_team\_location}.} 
To encode type-class neighbors $(r_{\textit{class}}, c)$, similar to the input representations of BERT \citep{Devlin_2019}, we build the input  sequence $H=(\texttt{[CLS]},r_{\textit{class}},c)$, where \texttt{[CLS]} is a special token, and for each element $h_i$ in  ${H}$, we construct its input vector representation $\textbf{h}_i$ as:
\begin{equation*}
\textbf{h}_i = \textbf{h}_i^{word} + \textbf{h}_i^{pos}
\end{equation*}
\vic{$\textbf{h}_i^{word}$ and $\textbf{h}_i^{pos}$ are  randomly initialized word  and position embeddings of $r_{\textit{class}}$ or $c$. We apply a local transformer   to each type-class neighbor sequence to  model the interaction between the class relations and types of an entity. The output embedding corresponding to  $\texttt{[CLS]}$, denoted as $\textbf{H}^{cls} \in\mathbb{R}^{d \times 1}$, is then used to infer  missing types of the target entity, where $d$ represents the dimension of the embedding. For an entity with $n$ type-class neighbors,  they are denoted as $[\textbf{H}_1^{cls},\textbf{H}_2^{cls},...,\textbf{H}_{n}^{cls}]$ after the local transformer representation.}

\vic{Type-class neighbors are not capable to fully capture the structural information within the KG. To alleviate this problem, we also consider relational neighbors. As for type-class neighbors, to encode the relational neighbors $(r,f)$ of an entity, we build a sequence $Q=(\texttt{[CLS]},r,f)$, and  aggregate the word and position embeddings and further apply a local transformer. The output embedding of $\texttt{[CLS]}$ is denoted as $\textbf{Q}^{cls} \in\mathbb{R}^{d \times 1}$, and  for an entity with $m$ relational neighbors, they are represented as $[\textbf{Q}_1^{cls},\textbf{Q}_2^{cls},...,\textbf{Q}_{m}^{cls}]$ after the local transformer representation.}

\vic{The local transformer mainly pays attention to a single existing neighbor at a time in the inference process, reducing the  interference between unrelated types. We perform a non-linear activation on   neighbors, and then perform a linear layer operation to unify the dimension to  the number of types, the final local transformer score $\textbf{S}^{loc}\in\mathbb{R}^{L \times (m+n)}$ is defined as: }
\begin{equation}
\label{equation_local_score}
 \textbf{W}Relu([\textbf{H}_1^{cls},\ldots,\textbf{H}_{n}^{cls}, \textbf{Q}_1^{cls},\ldots,\textbf{Q}_{m}^{cls}]) + b
\end{equation}
\vic{$\textbf{W} \in\mathbb{R}^{L \times d}$ and $b \in\mathbb{R}^{L}$ are the learnable parameters, where $L$ is the number of types. $[,]$ denotes the concatenation function, $\textbf{H}_i^{cls}\in\mathbb{R}^{d \times 1}$ and  $\textbf{Q}_j^{cls} \in\mathbb{R}^{d \times 1}$ respectively represent the \textit{i-th} and \textit{j-th} embedding of the type-class and relational neighbors after the transformer representation.}


\vicf{An important observation is that} the number of relations available vary from one KG to another. For instance, the YAGO43kET KG has substantially fewer relations than the FB15kET KG (cf. the dataset statistics in the Experiments Section), making  the  discrimination among relations in relational triples harder. To tackle this problem, for the YAGO43kET KG,  we semantically enrich the representation of relations by using the type-class membership information. 
Specifically, for a relational neighbor $(r,f)$ of an entity,  we use the types of $f$ belonging to a certain class to enhance the relation $r$ in the sequence $(\texttt{[CLS]},r,f)$ using the following steps:
\begin{enumerate}[itemsep=0.5ex, leftmargin=5mm]
\item \vic{ Let  $ \Gamma =\{(\textit{has\_type},c_1),$ $(\textit{has\_type},c_2),$ $\ldots,(\textit{has\_type},c_\ell)\}$ be  the set of all type neighbors of $f$. We replace $\Gamma$ with the set  $\Gamma'$ of  corresponding type-class neighbors: $\{(r_{\textit{class}_1},c_1),$ $(r_{\textit{class}_2},c_2),\ldots,((r_{\textit{class}_\ell},c_\ell)\}$, i.e., representing that $c_i$ is a member of  $\textit{class}_i$.}
%
%
\item \vic{Based on $r$ and $\Gamma'$, we construct a sequence $P=(r,r_{\textit{class}_1},c_1,r_{\textit{class}_2},c_2,$ $\ldots,r_{\textit{class}_\ell},c_\ell)$. For each element $p_i$ of  $P$,  we assign randomly initialized word and position embeddings to capture sequence order. We then apply a transformer to capture the interaction between tokens. The output token embeddings are denoted as $[\textbf{p}_0, \textbf{p}_1, \ldots ,\textbf{p}_\ell]$.}
\item \vic{For the output token embeddings, we use three different operations to obtain the final  representation of relation $r$: average, maximum, and minimum. For the YAGO43kET KG, we replace the word embedding $r$ in sequence $Q$ with  $\textbf{P}_{avg} = \sum_{i=0}^\ell \textbf{p}_i$, $\textbf{P}_{max} = \textit{Max}(\textbf{p}_i)$, or $\textbf{P}_{min} = \textit{Min}(\textbf{p}_i)$.} 
\end{enumerate}

\subsubsection{Global Transformer} \label{sec:global}
\vic{ The local transformer mechanism is suitable for types that can be inferred by looking at simple structures, and for which independently considering neighbors is thus enough. However, inferring `complex' types requires to capture the interaction between different neighbors of an entity.  For instance, if we would like to  infer that the entity \textit{Birmingham\_City\_L.F.C.} has type \textit{Women's\_football\_clubs\_in\_England}, we need to simultaneously consider different sources of information to support this, \vicf{such as} the type neighbor  (\textit{has\_type}, \textit{Association\_football\_clubs}) and   relational neighbor (\textit{isLocatedIn}, \textit{England}) of \textit{Birmingham\_City\_L.F.C.}, and that (\textit{playsFor}, \textit{Birmingham\_City\_L.F.C.}) and (\textit{hasGender}, \textit{female}) are relational neighbors of the entity \textit{Darla\_Hood}. To this aim, we introduce a global transformer module capturing the  interaction between type-class and relational neighbors by comprehensively representing them as the input of a transformer as follows:}

\begin{enumerate}[itemsep=0.2ex, leftmargin=5mm]
\item \vic{For a target entity $e$, we define the set $\Gamma'$ as done in Section~\ref{sec:local}. Further, let $\Xi = \{(r_1,f_1), \ldots, (r_m,f_m)\}$ denote the set of all relational neighbors of $e$.}  
\item \vic{We uniformly represent $\Gamma'$ and $\Xi$ as a single sequence $G=(\texttt{[CLS]}$ $r_{\textit{class}_1},c_1,r_{\textit{class}_2},c_2,$ $\ldots,r_{\textit{class}_\ell},c_n,$ $r_1,f_1, r_2,f_2, \ldots r_m,f_m)$.}
\item \vic{
For each element in the sequence $G$, we assign  randomly initialized word and position embeddings, and input it into a transformer. The output embedding of $\texttt{[CLS]}$ is denoted $\textbf{G}^{cls} \in\mathbb{R}^{d \times 1} $. Similar to Equation~\eqref{equation_local_score}, we define the prediction score $\textbf{S}^{glo} \in\mathbb{R}^{L \times 1}$  as $ \textbf{W}Relu([\textbf{G}^{cls}]) + b$.}
\end{enumerate}

\subsubsection{Context Transformer} \label{sec:context}
\vic{For complex types, the global transformer uniformly serializes the information about the neighbors of the target  entity. However, the neighbors of the target entity  are pairs, and this structural information might be useful for inference.  For instance,  to infer that  the entity \textit{Barack Obama} has type \textit{20th-century American writers}, we need to consider different aspects of its relational neighbors, e.g., the  neighbor  (\textit{bornIn}, \textit{Chicago}) focuses on the birthplace, while the neighbor (\textit{write}, \textit{A Promised Land}) is concerned with possible careers. The global transformer serialization of  pairs as a sequence  may lead to two problems: First, serializing neighbors disregards the structure of the graph. Second,  the importance of each element in the sequence is the same, and even elements that are not relevant for the inference will exchange information, e.g.,  \textit{bornIn} and  \textit{A Promised Land} in the example above. To realize a  differentiated aggregation  between different neighbor  pairs while preserving the graph structure, \vicf{we use a context transformer module as in~\citep{Chen_2021}.  Intuitively, given the output of the local transformer and  the $\texttt{[CLS]}$ embedding, the context transformer contextualizes the target entity with type-class and relational neighbors knowledge from its  neighborhood graph, details of the context transformer can be found in~\citep{Chen_2021}.} The output embedding of  $\texttt{[CLS]}$, denoted as $\textbf{C}^{cls} \in\mathbb{R}^{d \times 1}$, is used for the final entity type prediction, which is defined as $\textbf{S}^{ctx} = \textbf{W}Relu([\textbf{C}^{cls}]) + b$, where $\textbf{S}^{ctx} \in\mathbb{R}^{L \times 1}$. } 
\subsection{Pooling}
\label{section_3_3}
\vic{For an entity $e$, the local, global, and context transformers may generate multiple entity typing inference results. To address this, we adopt an exponentially weighted pooling method to aggregate prediction results~\citep{Pan_2021, Stergiou_2021},  formulated as follows:}
\begin{equation*}
\label{equation_pooling}
\textbf{S}_e = pool(\{\textbf{S}_0^{loc},\textbf{S}_1^{loc},...,\textbf{S}_{m+n-1}^{loc},\textbf{S}^{glo},\textbf{S}^{ctx}\})
\end{equation*}
 \vic{$\textbf{S}_e \in\mathbb{R}^{L}$ represents the relevance score between  $e$ and its types, and $n$ ($m$) is the number of type-class (relational) neighbors of $e$ respectively. For simplicity, we will omit the  identifiers (\textit{loc,glo,ctx}). We unify the numerical order of the output results of the local, global, and context transformers as follows:} \\
\begin{equation*}
\begin{split}
\label{equation_pooling_simplify}
\textbf{S}_e &{=} pool(\{\textbf{S}_0,\textbf{S}_1,...,\textbf{S}_{m+n-1},\textbf{S}_{m+n},\textbf{S}_{m+n+1}\}) \\ 
&=\sum_{i=0}^{m+n+1} w_i\textbf{S}_i,\,\,
w_i=  \frac{{\rm exp} \, \alpha\textbf{S}_i}
{\sum_{j=0}^{m+n+1} {\rm exp} \, \alpha\textbf{S}_j} \textbf{S}_i
\end{split}
\end{equation*}
 \vic{We further apply a sigmoid function to $\textbf{S}_e$, denoted as $\textbf{s}_e=\sigma(\textbf{S}_e)$, to map the scores between 0 and 1, where the higher the value of $\textbf{s}_{e,k}$ of $\textbf{s}_e$, the  more likely is $e$ to have type $k$.}
 
\subsection{Optimization Strategy}
\label{section_3_4}
\vic{To train a model with positive sample score $\textbf{s}_{e,k}$ (representing that $(e,\textit{has\_type}, k)$ exists in  a KG) and negative sample score $\textbf{s}_{e,k}'$ (representing that  $(e,\textit{has\_type}, k)$ does not exist in KG), usually   binary cross-entropy (BCE) is used as the loss function. However, there may exist a serious false negative problem, i.e., some $(e,\textit{has\_type}, k)$ are valid, but they are missing in existing KGs. To overcome this problem,  false-negative aware loss functions (FNA) have been proposed~\citep{Pan_2021}. Basically, they assign lower weight to  negative samples with too high or too low relevance scores. We introduce a  steeper false-negative aware (SFNA) loss function which  gives more penalties to negative samples with too high or too low relevance scores. The negative sample score is defined as:}
\begin{equation*}
f(x)=\left\{
	\begin{array}{l}
	3\,x-2\,x^2, \quad x<=0.5\\
	x-2\,x^2+1,\quad x>0.5\\
	\end{array} \right.
\end{equation*}
\vic{ For the positive score $\textbf{s}_{e,k}$ and negative score $\textbf{s}_{e,k}'$, the SFNA loss is defined as follows:}
\begin{equation*}
\mathcal{L}=-\sum f(\textbf{s}_{e,k}'){\rm log}(1-\textbf{s}_{e,k}')-\sum {\rm log}(\textbf{s}_{e,k})
\end{equation*}

\section{Experiments}

\begin{table*}[!htp]
\renewcommand\arraystretch{1.1}
\setlength{\tabcolsep}{0.53em}
\centering
\small
\begin{tabular*}{\linewidth}{@{}cccccccccc@{}}
\hline
\multicolumn{1}{c|}{\textbf{Datasets}} & \multicolumn{4}{c|}{\textbf{FB15kET}} & \multicolumn{4}{c}{\textbf{YAGO43kET}}\\
\hline
\multicolumn{1}{c|}{\textbf{Metrics}} & \multicolumn{1}{c}{\textbf{MRR}} & \multicolumn{1}{c}{\textbf{Hit@1}} & \multicolumn{1}{c}{\textbf{Hit@3}} & \multicolumn{1}{c|}{\textbf{Hit@10}} & \multicolumn{1}{c}{\textbf{MRR}} & \multicolumn{1}{c}{\textbf{Hit@1}} & \multicolumn{1}{c}{\textbf{Hit@3}} & \multicolumn{1}{c}{\textbf{Hit@10}} \\
\hline
\multicolumn{10}{c}{\textit{Embedding-based methods}} \\
\hline
\multicolumn{1}{l|}{ETE~\citep{Moon_2017}$^\lozenge$} &0.500  &0.385  &0.553  &\multicolumn{1}{c|}{0.719}  &0.230  &0.137  &0.263  &0.422 \\
\multicolumn{1}{l|}{ConnectE~\citep{Zhao_2020}$^\lozenge$} &0.590  &0.496  &0.643  &\multicolumn{1}{c|}{0.799}  &0.280  &0.160  &0.309  &0.479 \\
\multicolumn{1}{l|}{CORE-RotatE~\citep{Ge_2021}$^\lozenge$} &0.600  &0.493  &0.653  &\multicolumn{1}{c|}{0.811}  &0.320  &0.230  &0.366  &0.510 \\
\multicolumn{1}{l|}{CORE-ComplEx~\citep{Ge_2021}$^\lozenge$} &0.600  &0.489  &0.663  &\multicolumn{1}{c|}{0.816}  &0.350  &0.242  &0.392  &0.550 \\
\multicolumn{1}{l|}{AttEt~\citep{Zhuo_2022}$^\lozenge$} &0.620  &0.517  &0.677  &\multicolumn{1}{c|}{0.821}  &0.350  &0.244  &0.413  &0.565 \\
\multicolumn{1}{l|}{CET-BCE~\citep{Pan_2021}$^\lozenge$} &0.682  &0.593  &0.733  &\multicolumn{1}{c|}{0.852}  &0.472  &0.362  &0.540  &0.669 \\
\multicolumn{1}{l|}{CET-FNA~\citep{Pan_2021}$^\lozenge$} &0.697  &0.613  &0.745  &\multicolumn{1}{c|}{0.856}  &0.503  &0.398  &\uline{0.567}  &\textbf{0.696} \\
\hline
\multicolumn{10}{c}{\textit{GCN-based methods}} \\
\hline
\multicolumn{1}{l|}{HMGCN~\citep{Jin_2019}$^\blacklozenge$} &0.510  &0.390  &0.548  &\multicolumn{1}{c|}{0.724}  &0.250  &0.142  &0.273  &0.437 \\
\multicolumn{1}{l|}{ConnectE-MRGAT~\citep{Zhao_2022}$^\lozenge$} &0.630  &0.562  &0.662  &\multicolumn{1}{c|}{0.804}  &0.320  &0.243  &0.343  &0.482 \\
\multicolumn{1}{l|}{RACE2T~\citep{Zou_2022}$^\lozenge$} &0.640  &0.561  &0.689  &\multicolumn{1}{c|}{0.817}  &0.340  &0.248  &0.376  &0.523 \\
\multicolumn{1}{l|}{CompGCN-BCE~\citep{Vashishth_2020}$^\blacklozenge$} &0.657  &0.568  &0.704  &\multicolumn{1}{c|}{0.833}  &0.357  &0.274  &0.384  &0.520 \\
\multicolumn{1}{l|}{CompGCN-FNA~\citep{Vashishth_2020}$^\blacklozenge$} &0.665  &0.578  &0.712  &\multicolumn{1}{c|}{0.839}  &0.355  &0.274  &0.383  &0.513 \\
\multicolumn{1}{l|}{RGCN-BCE~\citep{Pan_2021}$^\lozenge$} &0.662  &0.571  &0.711  &\multicolumn{1}{c|}{0.836}  &0.357  &0.266  &0.392  &0.533 \\
\multicolumn{1}{l|}{RGCN-FNA~\citep{Pan_2021}$^\lozenge$} &0.679  &0.597  &0.722  &\multicolumn{1}{c|}{0.843}  &0.372  &0.281  &0.409  &0.549 \\
\hline
\multicolumn{10}{c}{\textit{Transformer-based methods}} \\
\hline
\multicolumn{1}{l|}{CoKE \citep{Wang_2019}$^\blacklozenge$} &0.465  &0.379  &0.510  &\multicolumn{1}{c|}{0.624}  &0.344  &0.244  &0.387  &0.542 \\
\multicolumn{1}{l|}{HittER \citep{Chen_2021}$^\blacklozenge$} &0.422  &0.333  &0.466  &\multicolumn{1}{c|}{0.588}  &0.240  &0.163  &0.259  &0.390 \\
\hline
\multicolumn{1}{c|}{TET-BCE} &0.699  &0.615  &0.748  &\multicolumn{1}{c|}{0.862}  &0.492  &0.385  &0.554  &0.684 \\
\multicolumn{1}{c|}{TET-FNA} &0.701  &0.608  &\uline{0.761}  &\multicolumn{1}{c|}{\textbf{0.873}}  &\uline{0.508}  &\uline{0.405}  &\uline{0.567}  &\textbf{0.696} \\
\multicolumn{1}{c|}{TET-SFNA-no-class} &\uline{0.706}  &\uline{0.626} &0.749  &\multicolumn{1}{c|}{0.862}  &0.472  &0.375  &0.525  &0.654 \\
\multicolumn{1}{c|}{TET-SFNA} &\textbf{0.717}  &\textbf{0.638}  &\textbf{0.762}  &\multicolumn{1}{c|}{\uline{0.872}}  &\textbf{0.510}  &\textbf{0.408}  &\textbf{0.571}  &\uline{0.695} \\
\hline
\end{tabular*}
\caption{Evaluation of different models on FB15kET and YAGO43kET. 
$\lozenge$ results are from the  original papers. $\blacklozenge$ results are from our implementation of the corresponding models. TET-SFNA-no-class means that type-class neighbors were not used, and for YAGO43kET in addition no semantic enhancement on relations is used. }
\label{table_main_result}
\end{table*}

\vic{In this section, we  discuss the evaluation of  TET relative to twelve baselines on a wide array of entity typing benchmarks. We first describe datasets and baseline models (Sec.~\ref{section_4_1}). Then we discuss the experimental results (Sec.~\ref{section_4_3}). Finally, we present ablation study experiments (Sec.~\ref{section_4_4}).}
\begin{table}[!htp]
\renewcommand\arraystretch{1.1}
\setlength{\tabcolsep}{1.20em}
\centering
\small
\begin{tabular*}{\linewidth}{@{}ccc@{}}
\toprule
\textbf{Datasets} & \textbf{FB15kET} & \textbf{YAGO43kET} \\
\midrule
\multicolumn{1}{l}{\# Entities}   &14,951  &42,335 \\
\multicolumn{1}{l}{\# Relations}   &1,345  &37 \\
\multicolumn{1}{l}{\# Types}   &3,584  &45,182 \\
\multicolumn{1}{l}{\# Clusters}   &1,081  &6,583 \\
\multicolumn{1}{l}{\# Train.triples}   &483,142  &331,686 \\
\multicolumn{1}{l}{\# Train.tuples}   &136,618  &375,853 \\
\multicolumn{1}{l}{\# Valid}   &15,848  &43,111 \\
\multicolumn{1}{l}{\# Test}   &15,847  &43,119 \\
\bottomrule
\end{tabular*}
\caption{Statistics of Datasets.}
\label{table_datasets}
\end{table}
\subsection{Datasets and Baselines}
\label{section_4_1}
\paragraph{Datasets.} \vic{We evaluate our proposed TET model on two real-world knowledge graphs: FB15k \citep{Bordes_2013} and YAGO43k \citep{Moon_2017} which are the subgraphs of Freebase \citep{Bollacker_2008} and YAGO \citep{Suchanek_2007}, respectively.} 
\vic{FB15kET and YAGO43kET provide entity type instances which map entities from FB15k and YAGO43k to corresponding entity types.
For  fairness of the experimental comparison, we followed the standard train/test split as in the baselines. The basic statistics of all datasets are shown in Table~\ref{table_datasets}}.

\paragraph{Baselines.} \vic{
We compare TET with twelve state-of-the-art entity typing methods, and their variants. 
We consider the embedding-based models {ETE} \citep{Moon_2017}, {ConnectE} \citep{Zhao_2020}, {CORE} \citep{Ge_2021},  {AttEt} \citep{Zhuo_2022} and {CET} \citep{Pan_2021}. We consider the GCN-based models  {HMGCN} \citep{Jin_2019}, {RACE2T} \citep{Zou_2022}, {ConnectE-MRGAT} \citep{Zhao_2022}, {CompGCN} \citep{Vashishth_2020} and {RGCN} \citep{Pan_2021}. We also use as baselines two transformer-based methods for KGC, CoKE and HittER~\cite{Wang_2019,Chen_2021}}. It should be noted that \vic{in all reported experimental results, the \textbf{bold} numbers denote the \emph{best results} while the \uline{underlined} ones  the \emph{second best}.}

\subsection{Experimental Results} 
\label{section_4_3}
\vic{Table~\ref{table_main_result} presents the evaluation results of entity type prediction on FB15kET and YAGO43kET. We can observe that our model TET outperforms all baselines in terms of basically all metrics. These results demonstrate that transformers  more effectively encode the neighbor information of an entity. Specifically, when using the BCE and FNA loss functions, TET meets or exceeds the CET model (the best performing baseline). By using the SFNA loss function, we can get further performance improvement, especially in the MRR and Hit@1 metrics on FB15kET.} 
\vic{Furthermore, TET has different gains compared to CET  with respect to the Hit metrics. The improvement on Hit@1 is higher than on Hit@3 and Hit@10 because by using three different transformer modules TET can encode the neighborhood information of an  entity at three different levels of granularity.}
\vic{Further, if we do not use type-class neighbors and for the YAGO43kET dataset the type-class enrichment on relations is not present (TET-SNFA-no-class), we note that the performance of TET on the YAGO43kET dataset decreases  considerably. Intuitively,  the decrease on the YAGO43kE is larger  than on FB15k because  the graph structure of YAGO43k is sparser, has fewer relations, and a large number of types, making the semantic  type-class knowledge crucial.} 

\begin{table*}[!htp]
\renewcommand\arraystretch{1.1}
\setlength{\tabcolsep}{0.71em} 
\centering
\small
\begin{tabular*}{\linewidth}{@{}ccccccccccc@{}}
\toprule
\multicolumn{3}{c}{\textbf{Models}} & \multicolumn{4}{c}{\textbf{FB15kET}} & \multicolumn{4}{c}{\textbf{YAGO43kET}}\\
\midrule
\multicolumn{1}{c}{\textbf{Global}} & \multicolumn{1}{c}{\textbf{Local}}&  \multicolumn{1}{c|}{\textbf{Context}} & \multicolumn{1}{c}{\textbf{MRR}} & \multicolumn{1}{c}{\textbf{Hit@1}} & \multicolumn{1}{c}{\textbf{Hit@3}} & \multicolumn{1}{c|}{\textbf{Hit@10}} & \multicolumn{1}{c}{\textbf{MRR}} & \multicolumn{1}{c}{\textbf{Hit@1}} & \multicolumn{1}{c}{\textbf{Hit@3}} & \multicolumn{1}{c}{\textbf{Hit@10}} \\
\midrule
$\surd$ &$\surd$ &\multicolumn{1}{c|}{$\surd$}   &\textbf{0.717}  &\textbf{0.638}  &\textbf{0.762}  &\multicolumn{1}{c|}{\textbf{0.872}}  &\textbf{0.510}  &\textbf{0.408}  &\textbf{0.571}  &\uline{0.695} \\
$\surd$ &$\surd$ &\multicolumn{1}{c|}{} &\uline{0.713}  &\uline{0.632}  &\uline{0.759}  &\multicolumn{1}{c|}{\uline{0.871}} &0.503  &0.401  &0.561  &0.690 \\
$\surd$ & &\multicolumn{1}{c|}{$\surd$} &0.664  &0.578  &0.711  &\multicolumn{1}{c|}{0.829} &0.369  &0.289  &0.397  &0.524 \\
&$\surd$ &\multicolumn{1}{c|}{$\surd$} &0.700  &0.614  &0.752  &\multicolumn{1}{c|}{0.864} &\uline{0.509}  &\uline{0.407}  &\uline{0.568}  &\textbf{0.697} \\
\midrule[0.03pt]
$\surd$ & &\multicolumn{1}{c|}{} &0.660  &0.578  &0.702  &\multicolumn{1}{c|}{0.824} &0.365  &0.286  &0.392  &0.517 \\
&$\surd$ &\multicolumn{1}{c|}{} &0.684  &0.596  &0.732  &\multicolumn{1}{c|}{0.859} &0.494  &0.387  &0.555  &0.690 \\
& &\multicolumn{1}{c|}{$\surd$} &0.641  &0.554  &0.686  &\multicolumn{1}{c|}{0.817} &0.353  &0.280  &0.375  &0.493 \\
\bottomrule
\end{tabular*}
\caption{Evaluation of ablation study with different transformer modules combinations.} 
\label{table_ablation_global_local_context}
\end{table*}

\subsection{Ablation Studies} 
\label{section_4_4}
\vic{To verify the impact of each TET model component on the performance, we conduct  ablation studies on FB15kET and YAGO43kET. In particular we look at the effect of: a) different transformer modules,  Table~\ref{table_ablation_global_local_context}; b) different neighbor content,  Table~\ref{table_ablation_neighbor_context}; c) different integration methods on YAGO43kET,  Table~\ref{table_ablation_with_different_integration}; d)  different dropping rates,  Table~\ref{table_ablation_study_with_different_dropping rates}; e) the number of hops,  Table~\ref{table_ablation_with_different_hop_paths}.}

\paragraph{Effect of Transformer.} 
\vic{The local transformer by itself performs better than the global one by itself. This indicates that considering independently the neighbors of an entity can reduce  interference between unrelated types. By combining  the global and context transformer,  more complex types can be inferred from the token and graph structure level,  achieving  state-of-the-art results.
} \vicf{Note that both the global and context transformers deal with complex types, but the context one further takes into account the relevance of different neighbors while preservin the structure of the KG. As one can see from the results, for the used datasets, the global transformer is already doing most of the work, i.e., the combination of local and global transformers achieves almost the same result as when the context one is also incorporated. We believe that in datasets with a more complex structure the context transformer could play a more prominent role, we leave this line of research as future work. }
\paragraph{Effect of Neighbor Content.} 
\vic{We observe that the impact of relational neighbors is greater than that of type-class neighbors. Indeed, removing relational neighbors leads to a substantial performance degradation in YAGO43kET. When both of them are available, type-class neighbors might help relational ones to distinguish between relevant and irrelevant types for an inference.}


\begin{table}[!htp]
\renewcommand\arraystretch{1.1}
\setlength{\tabcolsep}{0.35em}
\centering
\small
\begin{tabular*}{\linewidth}{@{}cccccc@{}}
\hline
\multicolumn{6}{c}{\textbf{FB15kET}} \\
\hline
\multicolumn{1}{c}{\textbf{relational}}&  \multicolumn{1}{c}{\textbf{type-class}} & \multicolumn{1}{c}{\textbf{MRR}} & \multicolumn{1}{c}{\textbf{Hit@1}} & \multicolumn{1}{c}{\textbf{Hit@3}} & \multicolumn{1}{c}{\textbf{Hit@10}} \\
\hline
$\surd$ &\multicolumn{1}{c|}{$\surd$}   &\textbf{0.717}  &\textbf{0.638}  &\textbf{0.762}  &\multicolumn{1}{c}{\textbf{0.872}} \\
$\surd$ &\multicolumn{1}{c|}{}   &\uline{0.657}  &\uline{0.568}  &\uline{0.707}  &\multicolumn{1}{c}{0.833} \\
&\multicolumn{1}{c|}{$\surd$}   &0.654  &0.561  &0.705  &\multicolumn{1}{c}{\uline{0.839}} \\
\hline
\multicolumn{6}{c}{\textbf{YAGO43kET}} \\
\hline
$\surd$ &\multicolumn{1}{c|}{$\surd$}   &\textbf{0.510}  &\textbf{0.408}  &\textbf{0.571}  &\multicolumn{1}{c}{\textbf{0.695}} \\
$\surd$ &\multicolumn{1}{c|}{}   &\uline{0.467}  &\uline{0.372}  &\uline{0.518}  &\multicolumn{1}{c}{\uline{0.642}} \\
&\multicolumn{1}{c|}{$\surd$}   &0.373  &0.288  &0.405  &\multicolumn{1}{c}{0.535} \\
\hline
\end{tabular*}
\caption{Evaluation of ablation study with different neighbor content.} 
\label{table_ablation_neighbor_context}
\end{table}

\begin{table}[!htp]
\renewcommand\arraystretch{1.1}
\setlength{\tabcolsep}{0.35em}
\centering
\small
\begin{tabular*}{\linewidth}{@{}cccccccc@{}}
\toprule
\multicolumn{4}{c|}{\textbf{Models}} & \multicolumn{4}{c}{\textbf{YAGO43kET}} \\
\midrule
\multicolumn{1}{c}{\textbf{No}}& \multicolumn{1}{c}{\textbf{Avg}}&   \multicolumn{1}{c}{\textbf{Max}} &
\multicolumn{1}{c|}{\textbf{Min}} & \multicolumn{1}{c}{\textbf{MRR}} & \multicolumn{1}{c}{\textbf{Hit@1}} & \multicolumn{1}{c}{\textbf{Hit@3}} & \multicolumn{1}{c}{\textbf{Hit@10}} \\
\midrule
$\surd$& &  &\multicolumn{1}{c|}{}  &0.491  &0.385  &0.554  &0.684  \\
& $\surd$ & &\multicolumn{1}{c|}{}   &\textbf{0.510}  &\textbf{0.408}  &\textbf{0.571}  &\textbf{0.695}  \\
& & $\surd$ &\multicolumn{1}{c|}{}  &\uline{0.505}  &0.404  &\uline{0.564}  &0.688  \\
& &  &\multicolumn{1}{c|}{$\surd$}  &\uline{0.505}  &\uline{0.405}  &0.563  &\uline{0.691}  \\
\bottomrule
\end{tabular*}
\caption{Evaluation of ablation study with different integration methods. 
Note that, "No" means without performing type-class semantic enhancement on the relations.}
\label{table_ablation_with_different_integration}
\end{table}

\paragraph{Effect of Integration Methods.} \vic{YAGO43kET has a sparser graph structure, fewer types of relations and a large number of types. To tackle this problem, in Section~\ref{sec:local}, we have enriched the representations of relations in   relational neighbors with type-class knowledge. One can observe that the Avg operation outperforms Min and Max because the latter tend to discard  useful content.}

\begin{table}[!htp]
\renewcommand\arraystretch{1.1}
\setlength{\tabcolsep}{0.09em}
\centering
\small
\begin{tabular*}{\linewidth}{@{}ccccccc@{}}
\toprule
\multicolumn{1}{c|}{\textbf{Dropping Rates}} & \multicolumn{3}{c|}{\textbf{75\%}} &\multicolumn{3}{c}{\textbf{90\%}}\\
\toprule
\multicolumn{1}{c|}{\textbf{Models}}  & \multicolumn{1}{c}{\textbf{MRR}} & \multicolumn{1}{c}{\textbf{Hit@1}} & \multicolumn{1}{c|}{\textbf{Hit@3}} & \multicolumn{1}{c}{\textbf{MRR}} & \multicolumn{1}{c}{\textbf{Hit@1}} & \multicolumn{1}{c}{\textbf{Hit@3}} \\
\toprule
\multicolumn{1}{c|}{CompGCN} 
&0.648  &0.559  &\multicolumn{1}{c|}{0.697}
&0.633  &0.544  &\multicolumn{1}{c}{0.679}\\
\multicolumn{1}{c|}{RGCN} 
&0.648  &0.560  &\multicolumn{1}{c|}{0.694}
&0.626  &0.534  &\multicolumn{1}{c}{0.673}\\
\multicolumn{1}{c|}{CET} 
&0.670  &0.580  &\multicolumn{1}{c|}{\uline{0.721}}
&\uline{0.646}  &0.553  &\multicolumn{1}{c}{\uline{0.698}}\\
\multicolumn{1}{c|}{TET\_RSE}
&\uline{0.683}  &\uline{0.599}  &\multicolumn{1}{c|}{\textbf{0.733}}
&0.645  &\uline{0.555}  &\multicolumn{1}{c}{0.692}\\
\multicolumn{1}{c|}{TET} 
&\textbf{0.689}  &\textbf{0.606}  &\multicolumn{1}{c|}{\textbf{0.733}}
&\textbf{0.658}  &\textbf{0.574}  &\multicolumn{1}{c}{\textbf{0.701}}\\
\bottomrule
\end{tabular*}
\caption{Evaluation with different dropping rates on FB15kET.  TET\_RSE represents TET with  semantic enhancement on relations. 
}
\label{table_ablation_study_with_different_dropping rates}
\end{table}

\paragraph{Effect of Dropping Rates.} \vic{In real life KGs, many entities  have sparse relations with other entities. In particular, they have few relational neighbors but a large number of types, so for their inference  we lack  structural relational information. Indeed, in YAGO43kET about 4.73\% of its entities have  five times more types than relational neighbors~\citep{Zhuo_2022}. To further test the robustness of TET under relation-sparsity, we also conduct ablation experiments on FB15kET by randomly removing 25\%, 50\%, 75\%, and 90\% of the relational neighbors of entities. We find that with the continuous increase of the sparsity ratio, the performance of baselines decrease to varying degrees, but TET still achieves the best results under all sparsity conditions. We also consider TET with semantic enhancement on relations since by randomly dropping neighbors the number of relations might also be reduced. However, not enough relations are removed to have a positive effect. Another reason for not having positive effect is that  the number of types in FB15kET is substantially smaller than in  YAGO43kET. Table~\ref{table_ablation_study_with_different_dropping rates} shows results for 75\%, and 90\%, for missing results see appendix.}

\begin{table}[!htp]
\renewcommand\arraystretch{1.1}
\setlength{\tabcolsep}{0.30em}
\centering
\small
\begin{tabular*}{\linewidth}{@{}ccccccc@{}}
\toprule
\multicolumn{3}{c}{\textbf{Models}} & \multicolumn{4}{c}{\textbf{FB15kET}} \\
\midrule
\multicolumn{1}{c}{\textbf{1-hop}}&   \multicolumn{1}{c}{\textbf{2-hops}} &
\multicolumn{1}{c|}{\textbf{3-hops}} & \multicolumn{1}{c}{\textbf{MRR}} & \multicolumn{1}{c}{\textbf{Hit@1}} & \multicolumn{1}{c}{\textbf{Hit@3}} & \multicolumn{1}{c}{\textbf{Hit@10}} \\
\midrule
$\surd$   &   &   \multicolumn{1}{c|}{}  &\textbf{0.717}  &\textbf{0.638}  &\textbf{0.762}  &\textbf{0.872}  \\
&   $\surd$   &   \multicolumn{1}{c|}{}  &0.677  &0.592  &0.720  &0.844  \\
&   &   \multicolumn{1}{c|}{$\surd$}  &0.682  &0.597  &0.728  &0.850  \\
\midrule[0.03pt]
$\surd$&   $\surd$   &   \multicolumn{1}{c|}{}  &0.709  &0.626  &\uline{0.759}  &\uline{0.869}  \\
$\surd$&   &   \multicolumn{1}{c|}{$\surd$}  &\uline{0.710}  &\uline{0.630}  &0.756  &0.868  \\
&   $\surd$&   \multicolumn{1}{c|}{$\surd$}  &0.680  &0.598  &0.721  &0.845  \\
$\surd$&   $\surd$&   \multicolumn{1}{c|}{$\surd$}  &0.709  &\uline{0.630}  &0.754  &0.865  \\
\bottomrule
\end{tabular*}
\caption{Evaluation of ablation study with different number of hops on FB15kET.} 
\label{table_ablation_with_different_hop_paths}
\end{table}

\paragraph{Effect of Number of Hops.} \vic{For relational neighbors, TET only considers one-hop information i.e., only the information around their direct  neighbors. 
We also conduct an ablation study on the effect of using different number of hops. In principle multi-hop information could provide richer structural  knowledge, increasing the discrimination of relational neighbors. \vicf{ Indeed, a positive effect of multi-hop information has been witnessed  in several approaches to KGC.} However, our experimental results show that the noise introduced by intermediate  entities is more dominant than the additional knowledge  $n$-hop entities and relations provide.} \vicf{Intuitively, for KGC multi-hop information makes a difference as it  exploits the topological structure of the KG (i.e. how entities are related). However, in the input KG, types are not related between them and as our experiments show,  one can not lift the topological structure at the entity-level to the type one, explaining why there is no gain from considering multi-hop information. It would interesting to confirm this observation by using GCNs, which  more naturally capture multi-hop information.}

\section{Conclusions}
\vic{In this paper, we propose a novel transformer-based model for KGET which utilizes 
contextual information of entities  to infer  missing types for KGs with minimal schema information. TET has three modules allowing to encode local and global neighborhood information from different perspectives. We also enhance the representation of entities by using class membership knowledge of types. We experimentally showed the benefits of our model.}

\section{Limitations}
\vic{Our TET model currently suffers from two limitations. From the methodological viewpoint, a transformer mechanism  introduces more parameters than  embedding-based methods, bringing some computational burden and memory overhead, but they are tolerable. Also, there exist other important tasks related to types, e.g. fine-grained entity typing, aiming at  classifying entity mentions into  fine-grained semantic labels. TET is currently not appropriate for this kind of tasks.}

\section*{Acknowledgments}
This work has been supported by the National Natural Science Foundation of China (No.61936012), by the National Key Research and Development Program of China (No.2020AAA0106100), by the National Natural Science Foundation of China (No. 62076155),   by a  Leverhulme Trust Research Project Grant (RPG-2021-140), and by the Chang
Jiang Scholars Program (J2019032).

\bibliography{anthology,custom}
\bibliographystyle{acl_natbib}
\clearpage
\section*{Appendix}
\subsection*{A Details about Experiments}
In this section, we give  more experimental details and discuss the evaluation protocol.


\begin{table}[!hp]
\renewcommand\arraystretch{1.1}
\setlength{\tabcolsep}{0.30em}
\centering
\small
\begin{tabular*}{\linewidth}{@{}cc@{}}
\toprule
\multicolumn{1}{l}{\textbf{Parameter}} & \multicolumn{1}{c}{\textbf{\{FB15kET,\,\, YAGO43kET\}}} \\
\midrule
\multicolumn{1}{l}{\# Embedding dim}   &\multicolumn{1}{c}{\{100,\,\,\,100\}} \\
\multicolumn{1}{l}{\# Train or valid batch size}   &\multicolumn{1}{c}{\{128,\,\,\,128\}} \\
\multicolumn{1}{l}{\# Test batch size}   &\multicolumn{1}{c}{\{1,\,\,\,1\}} \\
\multicolumn{1}{l}{\# Learning rate}   &\multicolumn{1}{c}{\{0.001,\,\,\,0.001\}} \\
\multicolumn{1}{l}{\# Trm layers}   &\multicolumn{1}{c}{\{3,\,\,\,3\}} \\
\multicolumn{1}{l}{\# Trm hidden dim}   &\multicolumn{1}{c}{\{480,\,\,\,480\}} \\
\multicolumn{1}{l}{\# Trm heads}   &\multicolumn{1}{c}{\{4,\,\,\,4\}} \\
\multicolumn{1}{l}{\# Trm dropout rate}   &\multicolumn{1}{c}{\{0.2,\,\,\,0.2\}} \\
\multicolumn{1}{l}{\# Type neighbor sample size}   &\multicolumn{1}{c}{\{3,\,\,\,3\}} \\
\multicolumn{1}{l}{\# KG neighbor sample size}   &\multicolumn{1}{c}{\{7,\,\,\,8\}} \\
\multicolumn{1}{l}{\# Warmup epochs}   &\multicolumn{1}{c}{\{50,\,\,\,50\}} \\
\multicolumn{1}{l}{\# Valid epochs}   &\multicolumn{1}{c}{\{25,\,\,\,25\}} \\
\multicolumn{1}{l}{\# $\alpha$}   &\multicolumn{1}{c}{\{0.5,\,\,\,0.5\}} \\
\multicolumn{1}{l}{\# Epochs}   &\multicolumn{1}{c}{\{500,\,\,\,500\}} \\
\bottomrule
\end{tabular*}
\caption{The main hyperparameters of TET model in different datasets.  }
\label{table_hyperparameters}
\end{table}

\paragraph{Experimental Setting.}
\vic{Table~\ref{table_hyperparameters} shows the hyperparameter settings of the TET model on the FB15kET and YAGO43kET datasets. We use Adam \citep{Kingma_2015} as the optimizer, the hyperparameters are tuned according to the MRR on the validation set.  We only sample the entity type-class and relational neighbors during training and validation, but for testing we use all the neighbors of the entity for prediction, so the test batch size is set to 1. We adopted the warmup training strategy, keeping the initial learning rate 0.001 unchanged for the first 50 iterations, but after that we divided the learning rate by 5, and set the interval to be the current interval multiplied by 2. In Table~\ref{table_hyperparameters} "Trm" refers to the three transformer modules, we use the same number of layers, hidden dim, and heads.}

\paragraph{Evaluation Protocol.} \vic{For each test sample $(e,c)$, we first calculate the correlation score $\textbf{s}_e$ between  the entity $e$ and  type $c$, and then sort them in descending order. We report various metrics for evaluation, specifically, we adopt the filtered setting~\citep{Bordes_2013,Pan_2021}  for computing mean reciprocal rank (MRR), and the proportion of correct entities ranked in the top 1/3/10 (Hit@1, Hit@3, Hit@10).}

\subsection*{B Additional Results}
In this section, we discuss more ablation experimental results that are not included in the main part of the paper.
\paragraph{Effect of Dropping Rates: Number of Relational Neighbors.} In the
main body of the paper we discussed why we test the robustness of TET under relation-sparsity on FB15kET by randomly removing 25\%, 50\%, 75\%, and 90\% of the relational neighbors of entities. Due to space constraints we only presented the results for the two more extreme cases: 75\%, and 90\%. Table \ref{full_table_ablation_study_with_different_dropping rates} shows the missing results for 25\% and 50\%.

\begin{table}[!hp]
\renewcommand\arraystretch{1.1}
\setlength{\tabcolsep}{0.09em}
\centering
\small
\begin{tabular*}{\linewidth}{@{}ccccccc@{}}
\toprule
\multicolumn{1}{c|}{\textbf{Dropping Rates}} & \multicolumn{3}{c|}{\textbf{25\%}} & \multicolumn{3}{c}{\textbf{50\%}}\\
\toprule
\multicolumn{1}{c|}{\textbf{Models}}  & \multicolumn{1}{c}{\textbf{MRR}} & \multicolumn{1}{c}{\textbf{Hit@1}} & \multicolumn{1}{c|}{\textbf{Hit@3}} & \multicolumn{1}{c}{\textbf{MRR}} & \multicolumn{1}{c}{\textbf{Hit@1}} & \multicolumn{1}{c}{\textbf{Hit@3}}\\
\toprule
\multicolumn{1}{l|}{CompGCN} 
&0.661  &0.573  &\multicolumn{1}{c|}{0.705} 
&0.655  &0.565  &\multicolumn{1}{c}{0.702}\\
\multicolumn{1}{c|}{RGCN} 
&0.673  &0.590  &\multicolumn{1}{c|}{0.716} 
&0.667  &0.584  &\multicolumn{1}{c}{0.708}\\
\multicolumn{1}{c|}{CET} 
&0.697  &\uline{0.613}  &\multicolumn{1}{c|}{0.744} 
&0.687  &0.601  &\multicolumn{1}{c}{0.733}\\
\multicolumn{1}{c|}{TET\_RSE} 
&\uline{0.699}  &\uline{0.613}  &\multicolumn{1}{c|}{\uline{0.748}} 
&\uline{0.698}  &\uline{0.613}  &\multicolumn{1}{c}{\uline{0.749}}\\
\multicolumn{1}{c|}{TET} 
&\textbf{0.712}  &\textbf{0.631}  &\multicolumn{1}{c|}{\textbf{0.758}} 
&\textbf{0.705}  &\textbf{0.624}  &\multicolumn{1}{c}{\textbf{0.753}}\\
\bottomrule
\end{tabular*}
\caption{Evaluation with different dropping rates on FB15kET.  TET\_RSE represents TET with  semantic enhancement on relations. 
}
\label{full_table_ablation_study_with_different_dropping rates}
\end{table}
\begin{table}[!hp]
\renewcommand\arraystretch{1.1}
\setlength{\tabcolsep}{0.09em}
\centering
\small
\begin{tabular*}{\linewidth}{@{}ccccccc@{}}
\hline
\multicolumn{1}{c|}{\textbf{Dropping Rates}} & \multicolumn{3}{c|}{\textbf{25\%}} & \multicolumn{3}{c}{\textbf{50\%}}\\
\hline
\multicolumn{1}{c|}{\textbf{Models}}  & \multicolumn{1}{c}{\textbf{MRR}} & \multicolumn{1}{c}{\textbf{Hit@1}} & \multicolumn{1}{c|}{\textbf{Hit@3}} & \multicolumn{1}{c}{\textbf{MRR}} & \multicolumn{1}{c}{\textbf{Hit@1}} & \multicolumn{1}{c}{\textbf{Hit@3}}  \\
\hline
\multicolumn{1}{c|}{CompGCN} 
&0.664  &0.578  &\multicolumn{1}{c|}{0.708} 
&0.662  &0.574  &\multicolumn{1}{c}{0.708}\\
\multicolumn{1}{c|}{RGCN} 
&0.676  &0.593  &\multicolumn{1}{c|}{0.719} 
&0.673  &0.590  &\multicolumn{1}{c}{0.719}\\
\multicolumn{1}{c|}{CET} 
&0.699  &0.617  &\multicolumn{1}{c|}{0.743} 
&0.694  &0.610  &\multicolumn{1}{c}{0.742}\\
\multicolumn{1}{c|}{TET\_RSE} 
&0.708  &0.628  &\multicolumn{1}{c|}{0.754}
&\textbf{0.712}  &\textbf{0.634}  &\multicolumn{1}{c}{\uline{0.755}}\\
\multicolumn{1}{c|}{TET} 
&\textbf{0.711}  &\textbf{0.631}  &\multicolumn{1}{c|}{\textbf{0.756}} 
&\uline{0.710}  &\uline{0.630}  &\multicolumn{1}{c}{\textbf{0.757}}\\
\hline
\multicolumn{1}{c|}{\textbf{Dropping Rates}} & \multicolumn{3}{c|}{\textbf{75\%}} &\multicolumn{3}{c}{\textbf{90\%}}\\
\hline
\multicolumn{1}{c|}{\textbf{Models}}  & \multicolumn{1}{c}{\textbf{MRR}} & \multicolumn{1}{c}{\textbf{Hit@1}} & \multicolumn{1}{c|}{\textbf{Hit@3}} & \multicolumn{1}{c}{\textbf{MRR}} & \multicolumn{1}{c}{\textbf{Hit@1}} & \multicolumn{1}{c}{\textbf{Hit@3}} \\
\hline
\multicolumn{1}{c|}{CompGCN}
&0.653  &0.565  &\multicolumn{1}{c|}{0.699}
&0.637  &0.546  &\multicolumn{1}{c}{0.683}\\
\multicolumn{1}{c|}{RGCN}
&0.658  &0.573  &\multicolumn{1}{c|}{0.702}
&0.636  &0.548  &\multicolumn{1}{c}{0.681}\\
\multicolumn{1}{c|}{CET}
&0.675  &0.588  &\multicolumn{1}{c|}{0.721}
&0.653  &\uline{0.564}  &\multicolumn{1}{c}{0.700}\\
\multicolumn{1}{c|}{TET\_RSE} 
&\uline{0.687}  &\uline{0.604}  &\multicolumn{1}{c|}{\uline{0.733}}
&\uline{0.675}  &\textbf{0.591}  &\multicolumn{1}{c}{\uline{0.718}}\\
\multicolumn{1}{c|}{TET}
&\textbf{0.690}  &\textbf{0.608}  &\multicolumn{1}{c|}{\textbf{0.734}}
&\textbf{0.677}  &\textbf{0.591}  &\multicolumn{1}{c}{\textbf{0.722}}\\
\hline
\end{tabular*}
\caption{Evaluation with different dropping rates on relations on FB15kET. The TET\_RSE represents TET with semantic enhancement on relations. 
}
\label{table_ablation_study_with_different_dropping rates_for_relations}
\end{table}

\paragraph{Effect of Dropping Rates: Number of Relation Types.}
In Section~\ref{sec:local}, we enhanced relations with semantic knowledge for the YAGO43kE KG. The main reason for doing this only for YAGO43kE is that it contains only very few relation types  (cf. Table \ref{table_datasets}), making the discrimination among relation triples harder. As a first step towards having a better understanding of the interplay of the number of relation types and the number of types in a KG, we conduct an ablation study in which  on FB15kET we randomly remove 25\%, 50\%, 75\%, and 90\% of the relation types. The results in Table~\ref{table_ablation_study_with_different_dropping rates_for_relations} show that in this case enhancing relation types with semantic knowledge does not have a positive effect, unlike for YAGO43kE. We believe that the main reason behind this is that YAGO43kE does not only have very few relations, but also a very large number of types. To have a precise understanding, a dedicated deep analysis of the interplay of the number of types, the number of relation types, and other structural characteristics of KGs is required - it is out of the scope of this paper, but it is an interesting question for future work.

\end{document}